\RequirePackage{fix-cm}
\documentclass[pdftex,twocolumn]{svjour3}          
\smartqed  

\usepackage{bbding}
\usepackage{times}
\usepackage{epsfig}
\usepackage{graphicx}
\usepackage{caption}
\graphicspath{{figures/}}
\usepackage{amsmath}
\usepackage{amssymb}
\usepackage{autobreak}
\usepackage[colorlinks,linkcolor=red,anchorcolor=blue,citecolor=blue,CJKbookmarks=True]{hyperref}
\usepackage[utf8]{inputenc} 
\usepackage[T1]{fontenc}    
\usepackage{hyperref}       
\usepackage{booktabs}       
\usepackage{amsfonts}       
\usepackage{nicefrac}       
\usepackage{microtype}      
\usepackage{algorithm}
\usepackage{algpseudocode}
\usepackage{amsfonts,amssymb}
\usepackage{mathrsfs}
\usepackage{array}
\usepackage{bm}
\usepackage{multirow}
\usepackage{natbib}

\usepackage{times} 
\usepackage{helvet} 
\usepackage{courier} 
\usepackage{amsmath}
\usepackage{footnote}
\usepackage{multirow}
\usepackage{array}
\usepackage{algorithm}
\usepackage{algpseudocode}
\usepackage{amsfonts,amssymb}
\usepackage{booktabs}
\usepackage{xcolor}
\usepackage{verbatim}

\usepackage{ragged2e}

\usepackage{colortbl}
\usepackage{arydshln}
\usepackage{multirow}
\usepackage{multicol}
\usepackage{dashrule}
\usepackage{fancyhdr}
\usepackage{bbding}
\usepackage{url}

\usepackage{tabularx}
\usepackage{array}
\usepackage[misc]{ifsym}

\definecolor{mygray}{gray}{.9}
\definecolor{hidden-red}{RGB}{205, 44, 36}
\definecolor{hidden-blue}{RGB}{194,232,247}
\definecolor{hidden-orange}{RGB}{243,202,120}
\definecolor{hidden-green}{RGB}{34,139,34}
\definecolor{hidden-pink}{RGB}{255,245,247}
\definecolor{hidden-black}{RGB}{20,68,106}
\definecolor{LightRed}{rgb}{1,0.92,0.92}
\definecolor{LightOrange}{rgb}{1,0.95,0.88}
\definecolor{LightYellow}{rgb}{1.0,1.0,0.84}
\definecolor{LightGreen}{rgb}{0.9,1.0,0.88}
\definecolor{LightCyan}{rgb}{0.9,1,1}
\definecolor{LightBlue}{rgb}{0.9,0.94,1}
\definecolor{LightIndigo}{rgb}{0.92,0.9,1}
\definecolor{LightMagenta}{rgb}{0.96,0.86,1}
\definecolor{DirtyWhite}{rgb}{0.96,0.96,0.96}
\definecolor{DRed}{RGB}{255,0,0}
\journalname{International Journal of Computer Vision}
\begin{document}

\title{Zero-Shot Domain Generalization via  Domain Prompt Learning
}
\subtitle{}


\author{Shiming Chen\textsuperscript{1, \Letter}\thanks{Corresponding author: Shiming Chen. Email: gchenshiming@gmail.com} \and Shuhuang~Chen\textsuperscript{2} \and Salman Khan\textsuperscript{1,3} \and Fahad Shahbaz Khan\textsuperscript{1,4}\\
	\textsuperscript{1} Mohamed Bin Zayed University of Artificial Intelligence, UAE\\
	\textsuperscript{2} Huazhong University of Science and Technology, China\\
	\textsuperscript{3} Australian National University, Australia\\
	\textsuperscript{4}  Linköping  University, Sweden\\
	gchenshiming@gmail.com; \{salman.khan, fahad.khan\}@mbzuai.ac.ae\\ 
}


\date{Received: date / Accepted: date}

\maketitle

\begin{abstract}
	Domain generalization (DG) aims to improve
	the generalization of the model to any unseen domains. Targeting this goal, existing DG methods mainly focus on learning domain-invariant features or prompts from the given source domains. However, {these methods rely on source-domain data, and collecting and annotating large-scale single- or multi-source datasets is often infeasible, limiting the practical applicability of DG}. To tackle this challenge, we propose {\underline{Do}}main {\underline{P}}rompt {Learning} (DoP), {a method built} on a large {vision-language} model (e.g., CLIP) for zero-shot domain generalization, which recognizes unseen classes in unseen target domains without {access to source-domain data}. DoP essentially learns the target domain knowledge via prompt both in text and visual modalities to overcome the domain shift in text-visual modalities under the test-time adaptation formulation. Specifically, DoP first takes text domain learning (TDL) to simulate various distribution shifts by learning diverse style tokens implicitly in the text space with inter-style diversity loss and intra-style consistency loss.  Subsequently,  DoP employs visual domain learning (VDL) to further refine the target-domain-relevant style token generated by TDL by making use of the test visual data with self-supervised learning.  Extensive experiments on popular DG benchmark datasets (e.g., PACS, VLCS,  OfficeHome, and DomainNet)  and ImageNet show that our DoP consistently outperforms the state-of-the-art DG approaches. Codes are available at: \url{https://github.com/shiming-chen/DoP}.
	
\end{abstract}

\begin{figure}[h]
	\begin{center}
		\includegraphics[width=1\linewidth]{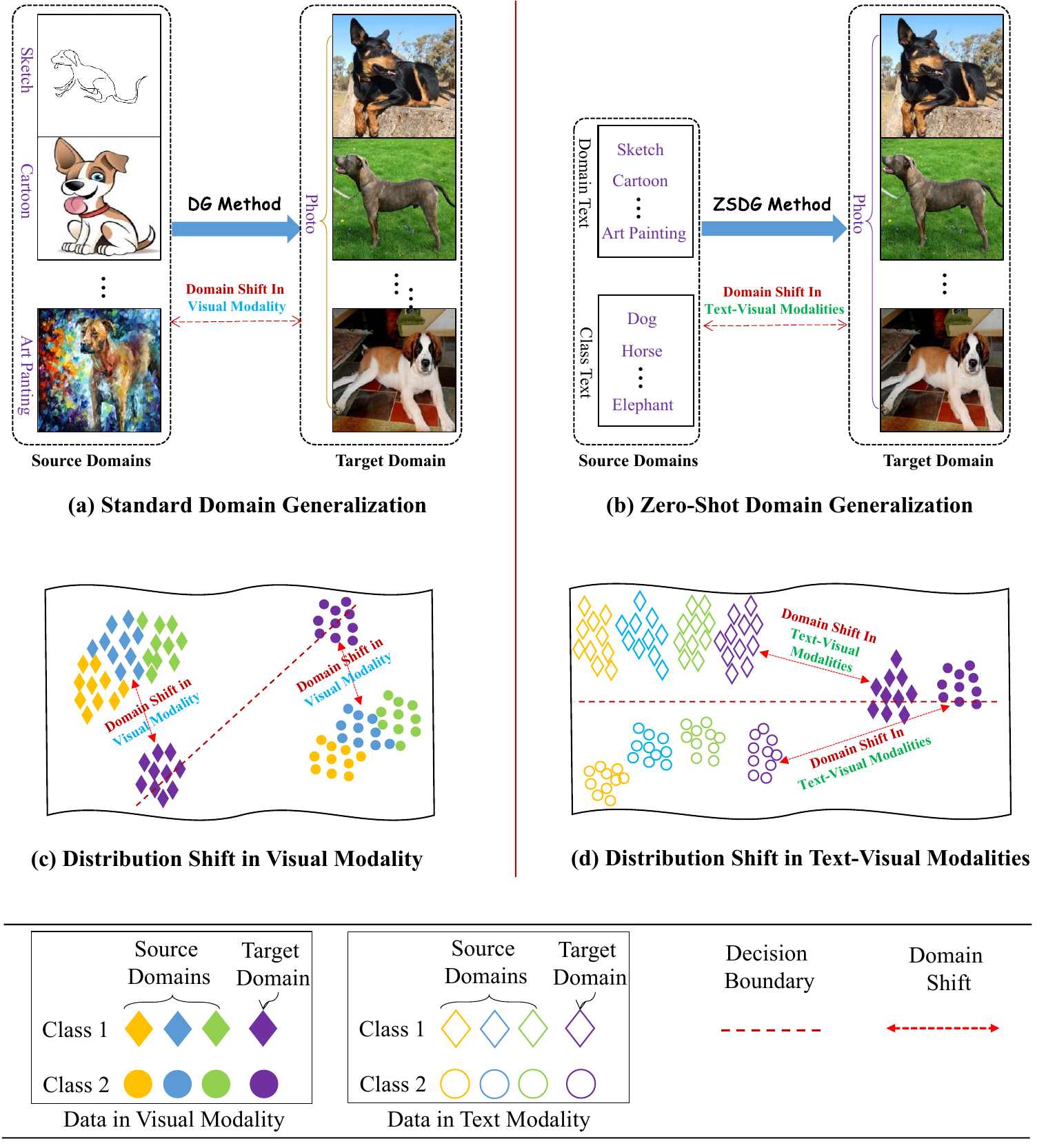}
		\caption{Standard Domain Generalization (DG)  \textit{vs} Zero-Shot Domain Generalization (ZSDG). (a)(c) Standard DG methods mainly learn the domain-invariant features or prompts from the given source domains to tackle \textit{domain shift in visual modality}. They are infeasible to collect and annotate large-scale source domain data for training, resulting in limited applications of DG in realistic scenarios. (b)(d) The ZSDG method recognizes unseen classes in the target/unseen domains without data in the source domain by overcoming \textit{domain shift in text-visual modalities}.}
		\label{fig:ZSDG}
	\end{center}
\end{figure}

\section{Introduction and Motivation}
Early machine learning models assume that training data and test data
are independent and identical distributions \cite{quinonero2008dataset, ChenWXPYZ021}, resulting in weak robust to significant distribution shifts between training and test distribution, e.g., style, backgrounds or day-to-night shifts.  
{To address this problem, domain} generalization (DG) {was introduced}\cite{BlanchardLS11,ZhouLQXL23}, {assuming that a} model is trained {on data from} one {or} multiple source domains and {generalizes well to} unseen target domains or styles. As shown in Fig.~\ref{fig:ZSDG}(a), given {source domains containing} images of dogs {in sketch}, cartoon, and {painting} styles, DG {aims to learn a} model that can recognize dogs {in} natural images or {photographs}. Enhancing the generalization ability of machine learning models is important in both industry and academic fields, DG has received much attention recently \cite{WangLLOQLCZY23}.

Standard DG methods aim to learn domain-invariant features or prompts from the given source domain data to tackle domain shift in the visual modality, as shown in Fig \ref{fig:ZSDG} (c). Early DG methods learn domain-invariant {features to improve model generalization to target domains} using various training strategies, such as data augmentation \cite{ChattopadhyayBH20,NamLPYY21,ZhouYHX20,NamLPYY21, NurielBW21},  cross-domain feature alignment methods \cite{GaninUAGLLML16, LiTGLLZT18, LiPWK18}, decomposing domain-specific and domain-invariant knowledge \cite{PiratlaNS20, ChattopadhyayBH20,LvLLZLWL22}, meta-learning  \cite{BalajiSC18,DouCKG19,Zhang0SG22}, and ensemble learning  \cite{ChaCLCPLP21, ZhouYQX21}. Although these methods achieve significant progress in small datasets, they cannot well generalize to large-scale datasets, e.g., different variants of ImageNet \cite{DengDSLL009}.

Recently, large {vision-language models (VLMs), such as} CLIP \cite{RadfordKHRGASAM21} and ALIGN \cite{JiaYXCPPLSLD21}, have been trained on large-scale {image–text data and have demonstrated strong} domain generalization and zero-shot learning {capabilities on downstream} tasks.
Accordingly, VLM reforms the DG  formulation and obtains new state-of-the-art generalization {performance} \cite{BaliahMSK23,AddepalliASB24}. There are two categories of methods to bring  VLM into DG, including prompt learning and fine-tuning VLM. For example, \cite{ZhouYLL22} introduces CoOp to explore the concept of prompt learning for automating prompt engineering via fine-tuning a few prompt tokens.  Subsequently,  he extends CoOp by further learning a lightweight neural network to generate an input-conditional token for each image and improve model generalization \cite{ZhouYL022}. In contrast, \cite{JiaTCCBHL22} proposed visual prompt tuning for fine-tuning VLM.  \cite{ChaLPC22}  re-formulated the DG objective using mutual information with the Oracle model to any possible domain. \cite{BaiZZLC24}  re-framed the prompt
learning framework from a generative perspective and propose a simple
yet efficient method for DG. 

We can find that the above methods access source domain data (i.e., visual data in some domains shared same classes in the target domain) for model optimization. However, {these methods face three major limitations} in realistic scenarios: i) it is unclear which source domains are ideal and how many source domains are sufficient for model learning since unseen domains are arbitrary in DG; ii) it is costly and  infeasible to collect and annotate large-scale source domain data for training; 
iii) Lots of classes cannot be collected in advance, especially for the new or scarce classes in the real world.  Naturally, it raises a new task, i.e., zero-shot domain generalization (ZSDG), which recognizes unseen classes in unseen domains without source domain data, as shown in Fig. \ref{fig:ZSDG} (b). 

The potential solution for ZSDG is directly making use of the zero-shot learning capacity of VLMs (e.g., CLIP) without accessing visual data of unseen classes in source data for training. Thus, the rest key task is how to discover the domain knowledge to improve the domain generalization of VLMs.  This raises the challenge of distribution shift in text-visual modalities in the unseen domain (as shown in Fig \ref{fig:ZSDG} (d)), which is more challenging than the distribution shift in the visual modality of standard DG. 

To tackle this challenge, our intuitive insight is to simultaneously discover target domain knowledge both in text and visual modalities under the test-time adaptation formulation \cite{SunWLMEH20,ShuNHYGAX22,Feng0LKZ23}. In the text modality, we {simulate diverse distribution shifts through prompts that synthesize diverse domain knowledge}. In the visual modality, we can explicitly discover the target domain knowledge via prompt by making use of test visual information with self-supervised learning during testing. More importantly, the domain knowledge learned in text modality can be further refined in the visual modality jointly.

In light of the above observations, we propose a novel \underline{Do}main \underline{P}rompt learning ( dubbed DoP) for zero-shot domain generalization. DoP consists of two key modules to learn the domain prompt, including text domain learning (TDL) and visual domain learning (VDL).  In the first stage, TDL simulates various distribution shifts via a prompt in the text space by synthesizing diverse domains knowledge, i.e., styles token generation using inter-style diversity loss and intra-style consistency loss and for optimization, and selects the most relevant style token (\textit{S*}) from the generated style tokens based on the test image. In the second stage, VDL refines the selected style token \textit{S*} using the test image and domain prompt (i.e., "\textit{A \textit{S*} style of a [class]}") via self-supervised learning. {Unlike PromptStyler~\cite{ChoNKYK23}, which uses all synthesized style features to construct a fixed domain-generalizable classifier, DoP treats the learned style tokens as candidate domain prompts and dynamically selects and visually refines the target-relevant token for each test image}. Finally, DoP takes the refined domain prompt to predict the classes of test images using CLIP prediction \cite{RadfordKHRGASAM21}. Extensive experiments demonstrate the effectiveness of our method.

The main contributions are summarized as follows: 
\begin{itemize}
	\item We introduce domain prompt learning, dubbed as DoP, which learns the domain knowledge both in text and visual spaces to tackle the domain shift in text-visual modalities of ZSDG.
	
	\item We deploy TDL  to discover the domain knowledge via a prompt in the text space using inter-style diversity loss and intra-style consistency loss. Meanwhile, we design VDL  to further refine the target-domain-relevant style token generated by TDL via self-supervised learning during testing.  Thus, DoP can implicitly and explicitly learn the domain knowledge with the text and visual supervision.
	
	\item Extensive experimental results on four {widely used} DG benchmarks (\textit{i.e.}, PACS, VLCS, OfficeHome, and DomainNet) and ImageNet,  demonstrate that our DoP learns the domain knowledge for effective ZSDG, thus leading to competitive {performance} over the current state-of-the-art methods.
\end{itemize}


\begin{figure*}[t]
	\begin{center}
		\includegraphics[width=1\linewidth]{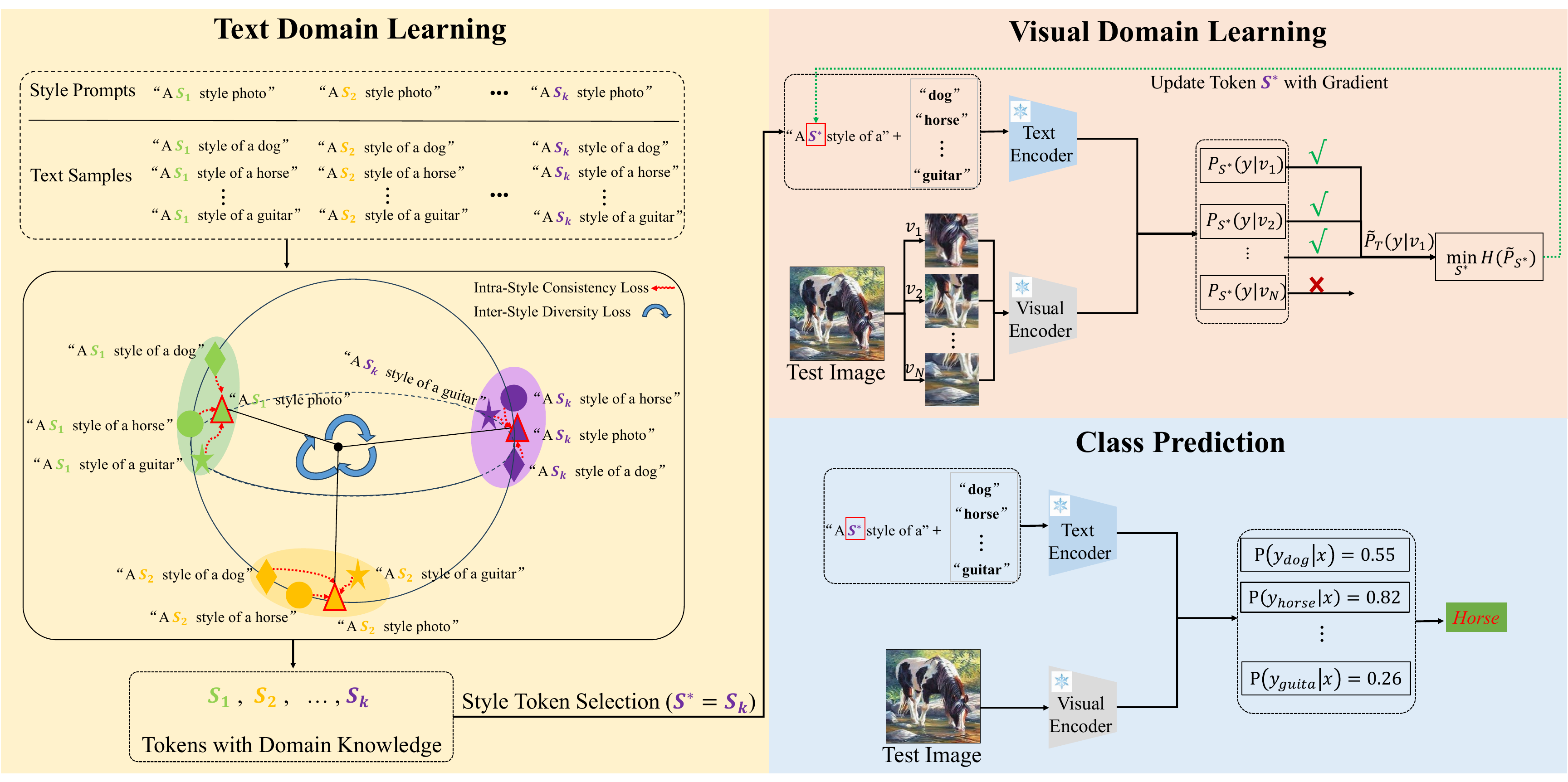}
		\\\vspace{-3mm}
		\caption{A schematic overview of DoP. DoP mainly consists of a text domain learning (TDL) module,  a visual domain learning (VDL) module, and a classifier.  Specifically, TDL first simulates various distribution shifts by learning diverse domain prompts implicitly in the text space with inter-style diversity loss and intra-style consistency loss and implicitly generates the style token $S^*$ that is relevant to the test image in target domains. Subsequently, VDL explicitly refines $S^*$ using the test visual data with self-supervised learning.  Finally, DoP adopts the refined domain prompt for class prediction following the standard CLIP. } 
		\label{fig:TDL}
	\end{center}
\end{figure*}

\section{Related Work}\label{sec2}

\noindent \textbf{Classical Domain Generalization.} Existing classical DG methods aim to learn domain-invariant feature representations to improve the generation of
the model on the target domain with different distributions from
the training data of the source domain. They focus on various training strategies on source domains, such as data augmentation,  cross-domain feature alignment,
decomposing domain-specific and domain-invariant knowledge, meta-learning,
and ensemble learning. 
Data augmentation methods \cite{ChattopadhyayBH20,NamLPYY21,ZhouYHX20,NamLPYY21, NurielBW21} {aim to increase} the diversity of {the training data and thereby enhance model generalization}.
For example, Zhou \cite{ZhouLL23,ZhouY0024} focuses on mixing the feature statistics of two random instances in a single forward pass during training.  Cross-domain feature alignment methods \cite{GaninUAGLLML16, LiTGLLZT18, LiPWK18}  align the distributions among different domains by adversarial training. Decomposing domain-specific and domain-invariant knowledge methods \cite{PiratlaNS20, ChattopadhyayBH20,LvLLZLWL22} disentangle the domain-specific and domain-invariant features and discard the domain-specific features for presenting the transferable features. Meta-learning methods \cite{BalajiSC18,DouCKG19,Zhang0SG22} leverage past knowledge to improve the learning of current tasks. For instance, MLDG \cite{LiYSH18} synthesizes meta-train and meta-test domains to simulate domain shift. Ensemble learning methods \cite{ChaCLCPLP21, ZhouYQX21} reduce the risk of overfitting by combining models learned from different domains or sampled from a series of episodes. 

These methods rely on utilizing the data of source domains for learning domain-invariant features. However, it is unclear which source domains are ideal and how many source domains are sufficient for model learning, since unseen domains are arbitrary in DG. Furthermore,
it is costly and sometimes even infeasible to collect and annotate large-scale source domain data for training. Most importantly, some classes are impossible to collect in advance, e.g., the new and scarce data.  Thus, it is necessary to develop new domain generalization methods, which enable the model to generalize to unseen classes in unseen domains.

\noindent \textbf{VLM-Based Domain Generalization.} 
The large vision-language (VLM)  is trained on great amounts of image-text pairs (e.g., CLIP was trained on 400M pairs) and has been applied to various downstream tasks, demonstrating their potential for remarkable domain generalization performance and zero-shot learning capabilities. Accordingly, VLM brings new insight for DG \cite{BaliahMSK23,ChoNKYK23,AddepalliASB24}, but the generation capacities are still hampered by the challenges of distribution shift \cite{ChenWZSMM24, BaiZZLC24} or the cross-dataset bias \cite{ChenWXPYZ021,ChenHHXSZYYS23}. {Fine-tuning VLMs is a promising approach for further improving their generalization on DG tasks}.  \cite{HuZQSG24} proposed a parameter-efficient group with orthogonal regularization to preserve the generalization ability of the pre-trained VLM and learn more diverse knowledge for DG. \cite{ShuG0W0L23} introduced CLIPPood to adapt CLIP models to DG via exploiting the semantic relations between classes from the text modality with the beta moving averaging optimization strategy. Another effective fine-tuning strategy is prompt learning, which only fine-tunes a few tokens to improve the generalization of VLM \cite{ZhouYLL22, ZhouYL022, JiaTCCBHL22}. {Different from multimodal domain generalization methods~\cite{dong2023simmmdg, fan2024cross, huang2025bridging} that learn domain-robust representations from multiple observed modalities, DoP focuses on mitigating the cross-modal distribution mismatch between textual prompts and visual representations in unseen domains without using source-domain images}.

{Most existing VLM-based DG methods require source-domain images for optimization, while PromptStyler~\cite{ChoNKYK23} learns diverse pseudo-style words without source data and uses the synthesized text features to train a fixed classifier. In contrast, DoP further performs test-image-conditioned style token selection and visual refinement for cross-modal domain prompt learning}. Thus, learning to discover the domain knowledge to improve the domain generalization capacity of existing VLM-based zero-shot learning methods is a good solution. Motivated by the test-time adaptation methods \cite{ShuNHYGAX22,Feng0LKZ23}, we design domain prompt learning to explicitly and implicitly learn target domain knowledge via prompt for zero-shot domain generalization.

\section{The Proposed Method}\label{sec3}

\noindent\textbf{Overview.} The framework of our domain prompt learning (DoP) is presented in Fig. \ref{fig:TDL}. Essentially, DoP implicitly and explicitly learns the domain knowledge of the target domain both in text and visual spaces to tackle the domain shift in text-visual modalities of ZSDG. DoP first takes text domain learning (TDL) to learn domain knowledge from the text space via inter-style diversity loss and intra-style consistency loss to generate a group of tokens of styles $\{S_1,  S_2, \cdots, S_k \}$, which is further selected one style token relevant to the target domain (e.g., $S^*$ ) based on the test data via token selection.  Subsequently, DoP employs visual domain learning (VDL)  to further refine $S^*$  using test data of the target domain by self-supervised learning. Finally, the domain prompt $\mathcal{P}^d$ (i.e., “\textit{A $S^*$ style of a [class]}”) learned by VDL is used for class prediction following the standard CLIP. {DoP addresses zero-shot domain generalization under} a test-time {prompt adaptation protocol. Specifically, ZSDG defines the considered data setting, in which no downstream source-domain images or target labels are available, whereas test-time adaptation describes how the domain prompt is refined using an unlabeled test instance. During VDL, only the selected style token is optimized, while the pretrained CLIP visual and text encoders remain frozen}.

\subsection{Text Domain Learning}\label{sec3.2}\vspace{-2mm}

Considering there are no source domain data that can be accessed, we can simulate various distribution shifts and generate the styles via a prompt in the text space. Targeting this goal, we first construct the text samples via prompts analogously to the visual data with both style and content information, i.e., “\textit{A $S_k$ style photo of a [class]}” (denoted as $\mathcal{P}_k^{style} \circ \mathcal{P}^{content}$).  Where $\{S_k\}_{k=1}^K$ are the style tokens {initialized from
randomly sampled style words}, and \textit{[class]} is the class name in the dataset. To help the model disentangle the style information from the text samples, we further construct the style prompt  “\textit{A $S_k$ style photo} ” (denoted as $\mathcal{P}_k^{style}$). Assuming we initiate $K$ style tokens and there are $M$ classes in the dataset, we can formulate $K\times M$ text samples and $K$ style prompts. We take the frozen CLIP text encoder (i.e., $E^T$) to extract their feature representations.

The first task in TDL is {to learn} tokens with various style concepts in the text space with two properties: i) {different style features should exhibit low pairwise correlation to reduce redundancy and improve the coverage of potential target domains}; ii) {text samples associated} with the same style concept should form compact clusters and remain distant {from samples associated with other styles}.

\noindent\textbf{Inter-Style Diversity Loss.} To enable TDL to learn diverse domain knowledge that can cover the target domains, we design inter-style diversity loss ($\mathcal{L}_{diversity}$) to optimize the style tokens of domain knowledge. For example, we fine-tuned the initialized style tokens $\{S_k\}_{k=1}^K$ by globally optimizing the styles features $\{E^T(\mathcal{P}_k^{style})\}_{k=1}^K$ produced by the $K$ style token, encouraging them to be orthogonal to each other. {Specifically, the cosine similarity between the $i$-th and
$j$-th style features} is formulated as:
\begin{gather}
	\label{Eq:style_cosine}
	z_{ij}^{\mathrm{style}}
	=
	\frac{
		E^T\left(\mathcal{P}_i^{\mathrm{style}}\right)
	}{
		\left\|
		E^T\left(\mathcal{P}_i^{\mathrm{style}}\right)
		\right\|_2
	}
	\cdot
	\frac{
		E^T\left(\mathcal{P}_j^{\mathrm{style}}\right)
	}{
		\left\|
		E^T\left(\mathcal{P}_j^{\mathrm{style}}\right)
		\right\|_2
	},
\end{gather}
{Based on the pairwise style similarity,
$\mathcal{L}_{\mathrm{diversity}}$ is formulated as}:
\begin{gather}
	\label{l_diversity}
    \begin{aligned}
	&\mathcal{L}_{\mathrm{diversity}}
	=
	\frac{K(K-1)}{2}
	\sum_{i=1}^{K}
	\sum_{j=1}^{K}
	\left|
	z_{ij}^{\mathrm{style}}
	\right|,\\
    &\text{where} \quad i \neq  j.
    \end{aligned}
\end{gather}
We can find that when $\mathcal{L}_{diversity}$ is minimized to be zero, it satisfies the orthogonality between all style features. {Here, orthogonality serves as a data-independent surrogate for reducing style redundancy rather than an assumption about real-world domain distributions}.

\noindent\textbf{Intra-Style Consistency Loss.}  Due to $\mathcal{L}_{diversity}$ only based on $K$ style prompts for learning, it is not enough to make style tokens with rich domain knowledge. To learn more reliable domain knowledge via prompts, we further take the text samples to discover domain knowledge. Specifically, we enforce the text samples with the same style token to be close to their corresponding style features and far away from the other style features using intra-style consistency loss ($\mathcal{L}_{consistency}$) for optimization. Thus, the contents of text samples are compacted and the style information is highlighted. As shown in Fig. \ref{fig:tsne}, the text samples with the same style token are clustered together.  That is, each text sample with $k$-th style token should have the highest similarity (e.g., cosine similarity score) with its corresponding style features and low similarity with other style features. For $t$-th class sample with $i$-th style (e.g., $\mathcal{P}_i^{style} \circ \mathcal{P}_t^{content}$), its cosine similarity score $z_{tij}$ with the  $j$-th style features formulated as:
\begin{gather}
	\label{Eq:l_consine}
	z_{t i j}=\frac{E^T\left(\mathcal{P}_i^{style }\circ \mathcal{P}_t^{content }  \right)}{\left\|E^T\left(  \mathcal{P}_i^{style } \circ \mathcal{P}_t^{content } \right)\right\|_2} \cdot \frac{E^T\left(\mathcal{P}_j^{style }\right)}{\left\|E^T\left(\mathcal{P}_j^{style }\right)\right\|_2},
\end{gather}
Based on the similarity, we formulate $\mathcal{L}_{consistency}$ for the $t$-th class with a contrastive loss: 
\begin{gather}
	\label{Eq:l_consistency}
	\mathcal{L}_{consistency }=-\frac{1}{K} \sum_{i=1}^K \log \left(\frac{\exp \left(z_{tii}\right)}{\sum_{j=1}^K \exp \left(z_{t i j}\right)}\right),
\end{gather}
Accordingly, $\mathcal{L}_{consistency}$ encourages each text sample feature to be {close} to its corresponding style {feature by suppressing} content information, {thereby enabling} the style tokens {to capture the shared style characteristics across classes. Different from PromptStyler~\cite{ChoNKYK23}, which sequentially learns style words with a class-centric content-consistency objective, our TDL jointly optimizes all $K$ style tokens using the inter-style diversity and intra-style consistency objectives while keeping the pretrained CLIP text encoder frozen}.

\begin{figure}[t]
	\begin{center}
		\includegraphics[width=0.95\linewidth]{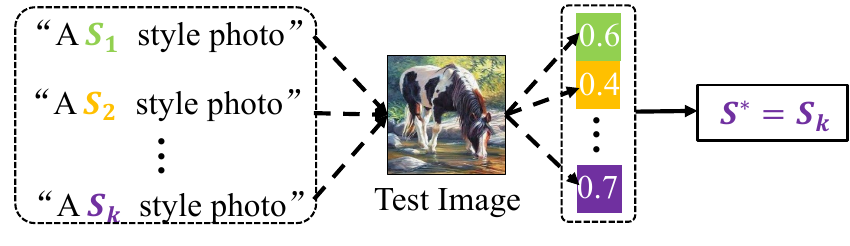}
		\\
		\caption{The process of style token selection.} 
		\label{fig:selection}
	\end{center}
\end{figure}

\noindent\textbf{Style Token Selection.} After learning the style tokens with various domain knowledge (i.e., $\{S_k\}_{k=1}^K$ ), we further select the most relevant style token for each test image. Thus,  the selected style token can be further refined based on the test image on visual space. The selection process is shown in Fig. \ref{fig:selection}. 
We first take the frozen visual encoder of CLIP (i.e., $E^V$) to extract the visual features of the test image. Then, we select the most relevant style token according to the cosine similarity scores between the test image $x$ and all learned style prompts $\{\mathcal{P}_k^{style}\}_{k=1}^K$:
\begin{gather}
	\begin{aligned}
		\label{Eq:selection}
		&S^*=\arg \max _{k}\{sim_k\}_{k=1}^K \\
		&  \text { where } \quad	sim_k=cosine(E^V(x), E^T(\mathcal{P}_k^{style})).
	\end{aligned}
\end{gather}
{This selection is performed independently for each test image using the frozen CLIP encoders without gradient updates, and only the highest-scoring style token $S^*$ is passed to VDL for subsequent visual refinement}.

\subsection{Visual Domain Learning}\label{sec3.3}
Due to the styles token learned by TDL  in the text space without the supervision of target data, they may not well capture the target domain knowledge and meet the issue of distribution shift in the visual modality. To further refine the style tokens in visual space, DoP takes VDL to fine-tune the domain prompt $\mathcal{P}^d$ (i.e., “\textit{A $S^*$ style of a [class]}”)  with the supervision of test image via test-time adaptation.  Different from test-time prompt tuning (TPT) \cite{ShuNHYGAX22}  that fine-tunes multiple {randomly} initialized tokens for learning context information in the test image, VDL only fine-tunes the style token $S*$ by discovering the domain knowledge from the test image. Accordingly, DoP can improve the generalization in any unseen domain.

\begin{figure*}[ht]
	\begin{center}
		\includegraphics[width=1\linewidth]{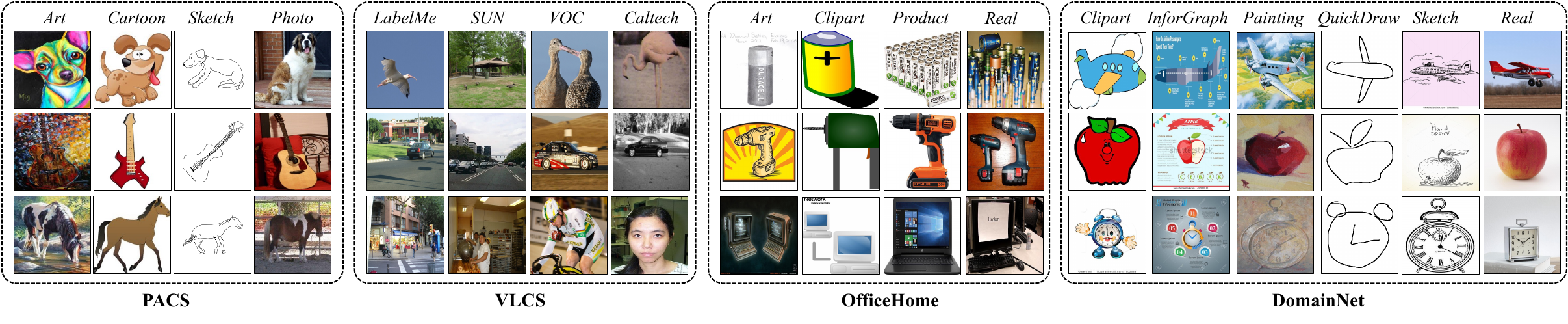}
		\\
		\caption{Visual diversity across domains in four widely used DG benchmarks. Representative samples from PACS~\cite{LiYSH17}, VLCS~\cite{FangXR13}, OfficeHome~\cite{VenkateswaraECP17}, and DomainNet~\cite{PengBXHSW19} illustrate substantial variations in rendering style, texture, background, and acquisition conditions, highlighting the difficulty of domain generalization.} 
		\label{fig:samples}
	\end{center}
\end{figure*}

For each test image $x$, VDL first {generates} $N$ augmented views, {denoted by ${v_1,v_2,\ldots,v_N}$, and uses the visual encoder $E^V$ to extract their features}. Then, it takes the selected style token $S^*$  learned by TDL and class names to construct $M$ domain prompts to extract the text features using $E^T$. Subsequently, VDL adopts the entropy to conduct self-supervision  and fine-tune  $S*$  based on the gradients: 
\begin{gather}
	\label{eq:entropy}
	\begin{aligned}
		& S^*=\arg \min _{S^*}-\sum_{i=1}^M \tilde{p}_{S^*}\left(y_i \mid x\right) \log \tilde{p}_{S^*}\left(y_i \mid x\right), \\
		&  \text { where } \quad \tilde{p}_{S^*}\left(y_i \mid x\right)=\frac{1}{N} \sum_{j=1}^N p_{S^*}\left(y_i \mid v_j\right).
	\end{aligned}
\end{gather}
Here, $\tilde{p}_{S^*}\left(y_i \mid x\right)$ is the averaged prediction of $x$, $p_{S^*}\left(y_i \mid v_j\right)$ is the class prediction probabilities produced by the domain prompt to $i$-th class and the $j$-th patches of $x$.

To reduce the noise from random augmentations, VDL filters out the patches of low confidence  (i.e., high-entropy) using a threshold  $\gamma$ with a confidence selection percentile $\rho$. Thus the  averaged prediction $\tilde{p}_{S^*}\left(y \mid x\right)$  can be rewritten as:
\begin{gather}
	\label{eq:filter}
	\tilde{p}_{S^*}\left(y_i \mid x\right)=\frac{1}{\rho N} \sum_{j=1}^N \leq \left[H\left(p_j\right) \leq \gamma \right] p_{S^*}\left(y_i \mid v_j\right),
\end{gather}
where $H(p_j)$ is the self-entropy of the prediction on the $j$-th patches. {For each test image, VDL independently refines the selected style token $S^*$ and discards the adapted token after prediction, preventing any information from being accumulated across test samples}.

\subsection{Class Prediction}\label{sec3.4}
After refining the style token $S^*$ in VDL,  our DoP performs class prediction following the standard CLIP. Differently, DoP takes the domain prompt $\mathcal{P}^d$  as text input of $E^T$. Thus, the $i$-th class prediction score of test sample $x$ is formulated as:
\begin{equation}
	p\left(y_i \mid x\right)=\frac{\exp \left(sim\left(E^T(\mathcal{P}_i^d) \cdot E^V(x)\right) \tau\right)}{\sum_{j=1}^M \exp \left(sim\left(E^T(\mathcal{P}_j^d) \cdot E^V(x)\right) \tau\right)}
\end{equation}
Where $\tau$ is the temperature of the softmax function and is set to 100 following CLIP. According to the prediction scores, DoP obtains the predicted class for  $x$.

\section{Experiments}\label{sec4}
\noindent\textbf{Datasets.} We evaluate our method on four {widely used} DG dataset benchmarks (e.g., PACS \cite{LiYSH17}, VLCS \cite{FangXR13}, OfficeHome \cite{VenkateswaraECP17}, and DomainNet \cite{PengBXHSW19}) and the large-scale dataset ImageNet (Including ImageNet-V2 \cite{RechtRSS19}, ImageNet-Sketch \cite{WangGLX19}, ImageNet-A \cite{HendrycksZBSS21} and
ImageNet-R \cite{HendrycksBMKWDD21}). Their statistics are summarized in Table \ref{Table:dataset}. Notably, most methods take the natural images in ImageNet \cite{DengDSLL009} for model training and evaluate them on the other four domains.  Since ZSDG does not require source data for training, we do not take the leave-one-domain-out cross-validation evaluation protocol \cite{PengBXHSW19} and take our DoP to test directly for all samples in datasets.

\begin{table}[ht]
	\centering
	\caption{Statistics of four widely used DG datasets (e.g., PACS, VLCS, OfficeHome,  and DomainNet ) and one large-scale DG dataset (i.e., ImageNet), including the numbers of domains/styles, classes, and images.} \label{Table:dataset}
	\setlength{\tabcolsep}{4.0mm}{\small
		\begin{tabular}{l|c|c|c}
			\hline       
			& Domains/Styles &     Classes       & Images \\   
			\hline
			PACS   & 4 &  7   & 9,991  \\
			VLCS   & 4& 5   & 10,729  \\
			OfficeHome  & 4  &  65    & 15,588  \\
			DomainNet  & 6  &  345    & 586,575 \\
			ImageNet  & 4  &  1000    & 98,389\\
			\hline
	\end{tabular}}
\end{table}

\noindent\textbf{Implementation Details.} We use {AdamW with a learning rate of $5\times10^{-3}$ for prompt optimization. TDL jointly optimizes all initialized style tokens} for 10 epochs, {whereas VDL independently updates only the selected token $S^*$ for each test image for two optimization rounds; the pretrained CLIP encoders remain frozen throughout both stages}. We empirically set the number of initialized style tokens $K=10/2/50/10/10$ for PACS, VLCS, OfficeHome, DomainNet, and ImageNet, respectively. We also experimentally set the confidence selection percentile $\rho$ to $\{0.1, 0.1, 0.2, 0.1, 0.1\}$ for  PACS, VLCS, OfficeHome, DomainNet, and ImageNet, respectively.

All experiments are performed on a single NVIDIA H100 graphic card with 80GB memory.
We use PyTorch\footnote{https://pytorch.org/} for the implementation of all experiments.

\begin{table*}[ht]
	\small
	\centering  
	\caption{Domain generalization results ~($\%$) of the state-of-the-art DG methods on PACS,  VLCS, OfficeHome, and DomainNet, including classical DG methods and VLM-based methods. Methods takes the source data for training marked with {\color{red}\CheckmarkBold}, otherwise, marked with {\color{red} \XSolidBrush}.}\label{table:sota}
	\resizebox{1.0\linewidth}{!}{\small
		\begin{tabular}{l|c|c|c|c|c|c}
			\hline
			Method &Source Data& PACS& VLCS &OfficeHome&  DomainNet &Average\\
			\hline
			&\multicolumn{6}{c}{\textit{ResNet-50 with pre-trained weights on ImageNet}}\\
			\cline{2-7}
			DANN~\cite{GaninUAGLLML16}&{\color{red}\CheckmarkBold}&83.6& 78.6& 65.9& 38.3& 66.6\\
			RSC~\cite{HuangWXH20}&{\color{red}\CheckmarkBold}&85.2& 77.1& 65.5& 38.9& 66.7\\
			SagNet~\cite{NamLPYY21}&{\color{red}\CheckmarkBold}& 86.3& 77.8& 68.1& 40.3& 68.1\\
			SelfReg~\cite{KimYPKL21}&{\color{red}\CheckmarkBold}&85.6& 77.8& 67.9& 42.8& 68.5\\
			GVRT~\cite{MinPKPK22}&{\color{red}\CheckmarkBold}&85.1 & 79.0 & 70.1& 44.1& 69.6\\
			MIRO~\cite{ChaLPC22}&{\color{red}\CheckmarkBold}&85.4 & 79.0 &\textbf{70.5} &44.3 &69.8\\
			SAGM~\cite{WangZLZ23}&{\color{red}\CheckmarkBold}&\textbf{86.6}&\textbf{80.0}& 70.1&  \textbf{45.0}&\textbf{70.4}\\
			\hline
			&\multicolumn{6}{c}{\textit{ResNet-50 with pre-trained weights from CLIP}}\\
			\cline{2-7}
			CAD \cite{RuanDM22} &{\color{red}\CheckmarkBold}&90.0& 81.2&70.5& 45.5& 71.8\\
			CoOp~\cite{ZhouYLL22}&{\color{red}\CheckmarkBold}&91.3& 81.4& 73.5& 49.7& 74.0\\
			CoCoOp~\cite{ZhouYL022} &{\color{red}\CheckmarkBold}&91.9& 81.8& 73.4& 49.7& 74.2\\
            PromptStyler~\cite{ChoNKYK23} &{\color{red}\XSolidBrush}&93.2 & \textbf{82.3} & 73.6 & 49.5 & 74.7\\
			\rowcolor{LightBlue}\textbf{DoP (Ours)} &{\color{red}\XSolidBrush}&\textbf{93.7}&	80.3& 	\textbf{74.7} &\textbf{50.7}&\textbf{74.9}\\
			\hline
			&\multicolumn{6}{c}{\textit{ViT-B / 16 with pre-trained weights from CLIP}}\\
			\cline{2-7}
			MIRO~\cite{ChaLPC22}&{\color{red}\CheckmarkBold}&95.6& 82.2 &82.5 &54.0& 78.6\\
			CoOp~\cite{ZhouYLL22}&{\color{red}\CheckmarkBold}&96.4& 80.8& 83.0& 59.5&79.9\\
			CoCoOp~\cite{ZhouYL022}&{\color{red}\CheckmarkBold}&96.7& 80.3& 83.4&  59.4& 80.0\\
			VPT~\cite{JiaTCCBHL22}&{\color{red}\CheckmarkBold}&96.9& 82.0& 83.2& 58.5&80.2\\
			MaPLe~\cite{KhattakR0KK23} &{\color{red}\CheckmarkBold}&96.5& 82.2& 83.4 & 59.5& 80.4 \\
			VL2V-SD~\cite{AddepalliASB24}&{\color{red}\CheckmarkBold}&94.9 &81.9 &85.7 & 59.4&80.5\\
			PEGO~\cite{HuZQSG24}&{\color{red}\CheckmarkBold}&96.5& 83.2& 84.2& 59.3& 80.8\\
			SPG~\cite{BaiZZLC24}&{\color{red}\CheckmarkBold}& 97.0& 82.4& 83.6&  60.1&80.8 \\
            PromptStyler~\cite{ChoNKYK23} &{\color{red}\XSolidBrush}&97.2 & 82.9 & 83.6 & 59.4 & 80.8\\
			\rowcolor{LightBlue}\textbf{DoP (Ours)} &{\color{red}\XSolidBrush}&\textbf{97.3} &	\textbf{84.3} &\textbf{84.7}&\textbf{61.4}&\textbf{82.0} \\
			\hline	
	\end{tabular} }
\end{table*}
\vspace{6mm}

\begin{table*}[ht]
	\small
	\centering  
	\caption{Domain generalization results ~($\%$) of the state-of-the-art methods on ImageNet. ViT-B/16 is used as a network backbone for fair comparison. Methods takes the source data for training marked with {\color{red}\CheckmarkBold}, otherwise, marked with {\color{red} \XSolidBrush}.} \label{table:zsdg-imagenet}
	\resizebox{1\linewidth}{!}{\small
		\begin{tabular}{l|c|c|c|c|c|c}
			\hline
			Method &Source Data&ImageNet-V2& ImageNet-R& ImageNet-S& ImageNet-A& Average\\
			\hline
			CLIP \cite{RadfordKHRGASAM21} &{\color{red}\XSolidBrush}&60.8& 74.0& 47.8 &46.1& 57.2\\
			CoOp \cite{ZhouYLL22}&{\color{red}\CheckmarkBold}& 64.2& 75.2& 47.9& 49.7& 59.3\\
			CoCoOp \cite{ZhouYL022}&{\color{red}\CheckmarkBold}&64.1& 76.2& 48.8& 50.6& 59.9\\
			TPT~\cite{ShuNHYGAX22}& {\color{red}\XSolidBrush}&64.3& 73.9& 46.4& 53.6& 59.5\\
			ProGrad~\cite{ZhuNHWZ23} &{\color{red}\CheckmarkBold}&\textbf{64.7}& 74.5& 47.9& 49.3& 59.1\\
			CuPL~\cite{PrattCLF23}& {\color{red}\XSolidBrush}& 63.3& 77.1& 48.8& 50.8&60.0\\
			ArGue \cite{TianZYZ24}&{\color{red}\CheckmarkBold}&64.6&76.6&48.9&50.9&60.3\\
			\rowcolor{LightBlue}\textbf{DoP (Ours)}&{\color{red}\XSolidBrush}&62.6&	\textbf{78.4}& \textbf{48.9} &\textbf{54.0}&\textbf{61.0}\\
			\hline
	\end{tabular} }
\end{table*}

\begin{table*}[ht]
	\small
	\centering  
	\caption{Zero-shot domain generalization results ~($\%$) of multiple methods on PACS,  VLCS, OfficeHome, and DomainNet. “DP” denotes the methods take “\textit{A random style of a [class]}”  as text prompt.}\label{table:zsdg-result}
	\resizebox{1.0\linewidth}{!}{\small
		\begin{tabular}{l|c|c|c|c|c}
			\hline
			Method & PACS& VLCS &OfficeHome&  DomainNet &Average\\
			\hline
			&\multicolumn{5}{c}{\textit{ResNet-50 with pre-trained weights from CLIP}}\\
			\cline{2-6}
			ZS-CLIP \cite{RadfordKHRGASAM21}&90.6&76.0& 68.6& 45.6& 70.2\\
			ZS-CLIP(DP) \cite{RadfordKHRGASAM21}&93.5 &79.8& 70.5& 	46.8  & 72.7 \\
			TPT~\cite{ShuNHYGAX22} &91.3 &81.7 & 72.9&49.1& 73.8\\
			TPT(DP)~\cite{ShuNHYGAX22}&93.5 &80.2&	71.2&49.1& 73.5 \\
            DPE~\cite{zhang2024dual} &91.8 & \textbf{82.0} & 73.5 & 47.3 & 73.7 \\
			\rowcolor{LightBlue}\textbf{DoP (Ours)}&\textbf{93.7}&	80.3& 	\textbf{74.7} &\textbf{50.7}&\textbf{74.9}\\
			\hline
			&\multicolumn{5}{c}{\textit{ViT-B / 16 with pre-trained weights from CLIP}}\\
			\cline{2-6}
			ZS-CLIP \cite{RadfordKHRGASAM21}&96.0 &82.8& 82.2&	57.5 &79.6\\
			ZS-CLIP(DP) \cite{RadfordKHRGASAM21}&96.7 &70.3& 80.7& 	57.3& 76.3  \\
			TPT~\cite{ShuNHYGAX22}&96.5 &83.3 &	83.8 &	59.6& 80.8\\
			TPT(DP)~\cite{ShuNHYGAX22}&96.6 &	70.4&81.8& 59.4 &77.1 \\
            PromptAlign~\cite{abdul2023align} & 93.5 & \textbf{85.3} & 81.1 & 55.8 & 78.9 \\
            DPE~\cite{zhang2024dual} &96.7 & 84.3 & 83.9 & 59.6 & 81.1 \\
			\rowcolor{LightBlue}\textbf{DoP (Ours)}&\textbf{97.3} &	84.3 &\textbf{84.7}&\textbf{61.4}&\textbf{82.0} \\
			\hline	
			&\multicolumn{5}{c}{\textit{ViT-L / 14 with pre-trained weights from CLIP}}\\
			\cline{2-6}
			ZS-CLIP \cite{RadfordKHRGASAM21}&98.5& 82.4&86.9& 64.0& 83.0\\
			ZS-CLIP(DP) \cite{RadfordKHRGASAM21}&97.7 &72.0&	87.2 &	63.0  &  80.0\\
			TPT~\cite{ShuNHYGAX22}&\textbf{98.8} &	\textbf{83.8 }&	88.6&65.4& \textbf{84.1}\\
			TPT(DP)~\cite{ShuNHYGAX22}&98.0 &74.5&	88.2&65.2& 81.5 \\
            DPE~\cite{zhang2024dual} &98.5 & 82.0 & 88.7 & 66.2 & 83.9 \\
			\rowcolor{LightBlue}\textbf{DoP (Ours)} &98.6 &78.9&\textbf{89.2}&\textbf{66.3}& 83.3\\
			\hline	
	\end{tabular} }
\end{table*}

\begin{table*}[ht]
	\small
	\centering
	\caption{Per-domain zero-shot domain generalization accuracy (\%)
	on PACS using CLIP-pretrained ResNet-50 and ViT-B/16 backbones.}\label{table:pacs-per-domain}
	\resizebox{0.8\linewidth}{!}{\small
		\begin{tabular}{l|c|c|c|c|c}
			\hline
			Method & Art& Cartoon &Photo& Sketch  &Average\\
			\hline
			&\multicolumn{5}{c}{\textit{ResNet-50 with pre-trained weights from CLIP}}\\
			\cline{2-6}
			ZS-CLIP \cite{RadfordKHRGASAM21}&88.9  & 94.4  & 99.3  & 79.8  & 90.6 \\
			TPT~\cite{ShuNHYGAX22} &92.5 & 93.0 &\textbf{99.4}&80.4& 91.3\\
            PromptStyler~\cite{ChoNKYK23} &93.7 & 94.7 & \textbf{99.4} & 84.9 & 93.2 \\
			\rowcolor{LightBlue}\textbf{DoP (Ours)}&\textbf{94.7} &\textbf{94.9} &\textbf{99.4} & \textbf{85.8}&\textbf{93.7}\\
			\hline
			&\multicolumn{5}{c}{\textit{ViT-B / 16 with pre-trained weights from CLIP}}\\
			\cline{2-6}
			ZS-CLIP \cite{RadfordKHRGASAM21}&96.8 & 98.8 & \textbf{99.9} & 88.3 & 96.0 \\
			TPT~\cite{ShuNHYGAX22}&97.8 &\textbf{99.2} &\textbf{99.9} &89.1& 96.5\\
            PromptStyler~\cite{ChoNKYK23} &97.6 & 99.1 & \textbf{99.9} & \textbf{92.3} & 97.2 \\
			\rowcolor{LightBlue}\textbf{DoP (Ours)}&\textbf{98.8} &99.0 &\textbf{99.9} &91.5 &\textbf{97.3} \\
			\hline	
	\end{tabular} }
\end{table*}

\begin{table}[t]
	\centering
	\caption{ Ablation studies for different components of our DoP on four datasets. “STS” denotes style token selection in TDL. We conduct these experiments using ViT-B/16 with pre-trained weights from CLIP.  } \label{Table:ablation-components}
	\resizebox{1.0\linewidth}{!}
	{\small
		\begin{tabular}{l|c|c|c|c|c}
			\hline
			Method & PACS& VLCS &OfficeHome&  DomainNet &Average\\
			\hline                           
			DoP w/o VDL &93.9&	77.3& 79.1& 52.7&75.8 \\   
			DoP w/o TDL &96.6 &70.4& 81.8& 59.4&77.1 \\     
			DoP w/o TDL(STS) &97.2 &	84.3 &83.5& 	60.7&81.4 \\ 
			DoP (Full) &\textbf{97.3 }&\textbf{84.3}& \textbf{84.7}&\textbf{61.4}&\textbf{82.0} \\   
			\hline 
		\end{tabular}
	}
\end{table}
{We compare DoP with} domain generalization methods using various network backbones, including classical DG methods (\textit{e.g.}, DANN~\cite{GaninUAGLLML16}, RSC~\cite{HuangWXH20}, SAGM~\cite{WangZLZ23}, SagNet~\cite{NamLPYY21}, and SelfReg~\cite{KimYPKL21}), VLM-based methods (\textit{e.g.}, CLIP~\cite{RadfordKHRGASAM21}, GVRT~\cite{MinPKPK22}, MIRO~\cite{ChaLPC22}, CoOp~\cite{ZhouYLL22}, CoCoOp~\cite{ZhouYL022}, TPT~\cite{ShuNHYGAX22}, VPT~\cite{JiaTCCBHL22}, MaPLe~\cite{KhattakR0KK23}, ProGrad~\cite{ZhuNHWZ23}, CuPL~\cite{PrattCLF23}, PEGO~\cite{HuZQSG24}, VL2V-SD~\cite{AddepalliASB24}, SPG~\cite{BaiZZLC24}, ArGue~\cite{TianZYZ24}, {PromptAlign~\cite{abdul2023align},  DPE~\cite{zhang2024dual}), and PromptStyler~\cite{ChoNKYK23}}.

\noindent\textbf{Comparison with Standard DG Methods.}
Table~\ref{table:sota} reports the results of standard DG methods {using different network backbones, including} ResNet-50 and ViT-B/16, {on four widely used} DG datasets. {Overall}, VLM-based methods consistently outperform classical DG methods, {demonstrating the strong generalization capability of pretrained VLMs} for DG tasks. {DoP achieves the best average accuracies of} 74.9\% and 82.0\% with ResNet-50 and ViT-B/16, respectively. {More} importantly, DoP {performs} zero-shot {domain generalization without requiring any downstream source-domain images, whereas most compared DG methods rely on source data for training. Compared with the PromptStyler~\cite{ChoNKYK23}, DoP improves the average accuracy by 0.2 and 1.2 percentage points with ResNet-50 and ViT-B/16, respectively, demonstrating the effectiveness of dynamically selecting and visually refining target-relevant domain prompts}.

We further report the generalization results on {the more challenging ImageNet variants} in Table~\ref{table:zsdg-imagenet}. DoP {achieves a competitive average accuracy} of 61.0\% {compared with strong VLM-based methods, including} ProGrad~\cite{ZhuNHWZ23}, CuPL~\cite{PrattCLF23}, and ArGue~\cite{TianZYZ24}. These results further demonstrate that DoP can effectively learn domain knowledge to {enhance the generalization capability of pretrained VLMs under diverse distribution shifts.}

\noindent\textbf{Comparison with Zero-Shot Learning Methods.}
Table~\ref{table:zsdg-result} reports the zero-shot domain generalization results of different methods using ResNet-50, ViT-B/16, and ViT-L/14. DoP achieves the best average accuracies of 74.9\% and 82.0\% with ResNet-50 and ViT-B/16, respectively. Compared with the recent test-time generalization methods, DoP outperforms DPE~\cite{zhang2024dual} by 1.2 percentage points with ResNet-50. With ViT-B/16, {DoP exceeds PromptAlign~\cite{abdul2023align} and DPE by 3.1 and 0.9 percentage points, respectively. Although PromptAlign and DPE perform better on VLCS in some settings, DoP achieves the best results on PACS, OfficeHome, and DomainNet, indicating more balanced generalization across different types of domain shift. With ViT-L/14, DoP achieves a competitive average accuracy of 83.3\%, compared with 84.1\% obtained by TPT~\cite{ShuNHYGAX22}}.

{We additionally incorporate explicit domain knowledge into the text prompts of ZS-CLIP and TPT, denoted as DP, using the template \textit{A random style of a [class]}.'' Although this direct prompting strategy yields slight improvements with ResNet-50, it generally degrades performance with ViT-B/16 and ViT-L/14, particularly on VLCS}.\footnote{VLCS mainly contains natural photographic images, as illustrated in Fig.~\ref{fig:samples}. Therefore, the conventional prompt \textit{A photo of a [class]}'' is better aligned with its visual content than a randomly specified style prompt.} {This observation suggests that directly inserting unlearned domain descriptions may disturb the pretrained text-visual alignment. In contrast, DoP learns diverse domain knowledge in the text space and further performs target-relevant prompt selection and visual refinement, thereby effectively mitigating the cross-modal distribution shift in ZSDG}.

\noindent\textbf{Per-Domain Analysis.}
{Table~\ref{table:pacs-per-domain} further reports the per-domain
performance on PACS. Compared with PromptStyler\cite{ChoNKYK23}, DoP consistently improves Art accuracy by 1.0 and 1.2 percentage points using ResNet-50 and ViT-B/16, respectively. With ResNet-50, DoP also improves Cartoon and Sketch by 0.2 and 0.9 percentage points, while maintaining the same Photo accuracy. With ViT-B/16, the differences on Photo and Cartoon are negligible, whereas Sketch accuracy decreases by 0.8 percentage points. Overall, DoP improves the average accuracy by 0.5 and 0.1 percentage points with the two backbones, respectively}. 

\begin{figure*}[ht]
	\begin{center}
		\includegraphics[width=1\linewidth]{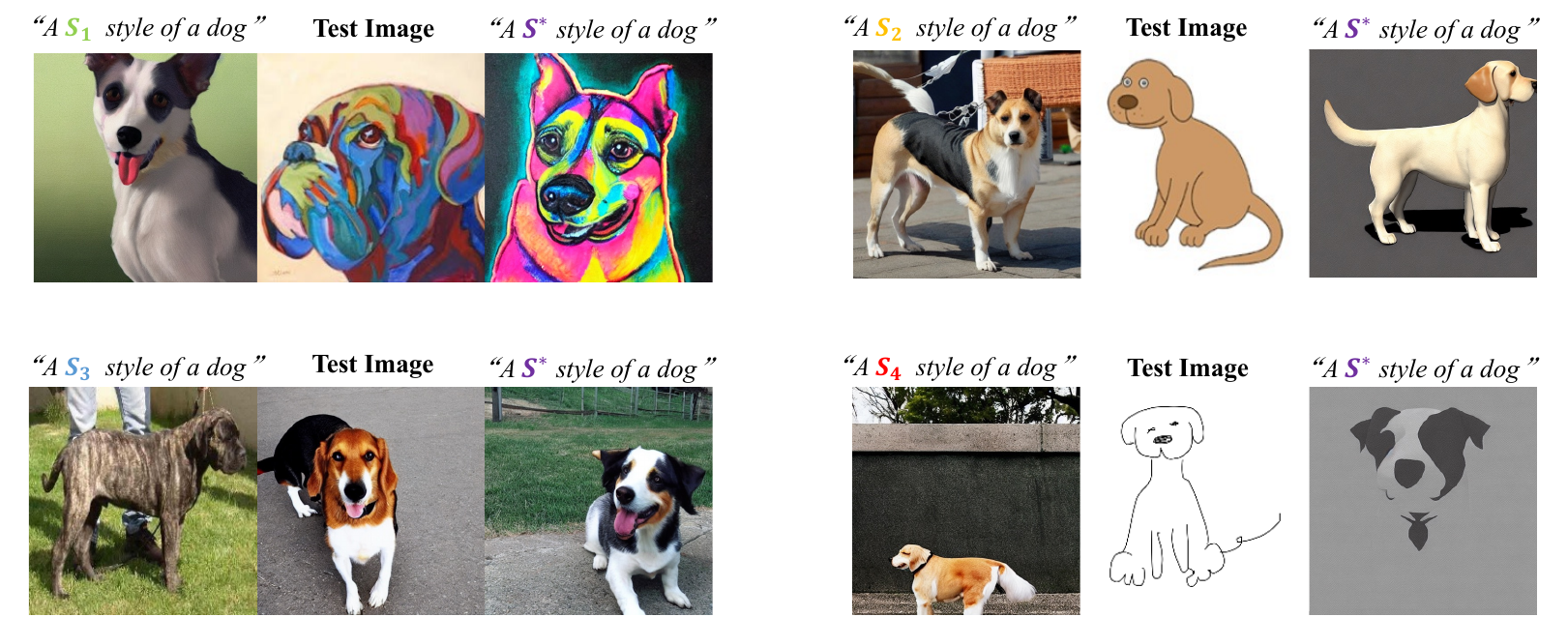}
		\\
		\caption{We take stable diffusion (SD) to synthesize images according to the domain prompts on PACS. For each group, the first image is generated by SD according to the domain prompt with randomly initialized style tokens, the second image is the real test image in the target domain, the third image is generated by SD according to the domain prompt with the style tokens learned by our DoP. Results show that DoP effectively discovers the domain knowledge with similar styles of test images in the target domain.} 
		\label{fig:style-visualization}
	\end{center}
\end{figure*}

\begin{figure*}
	\begin{center}
		\begin{tabular}{cccc}
			\includegraphics[width=0.23\linewidth]{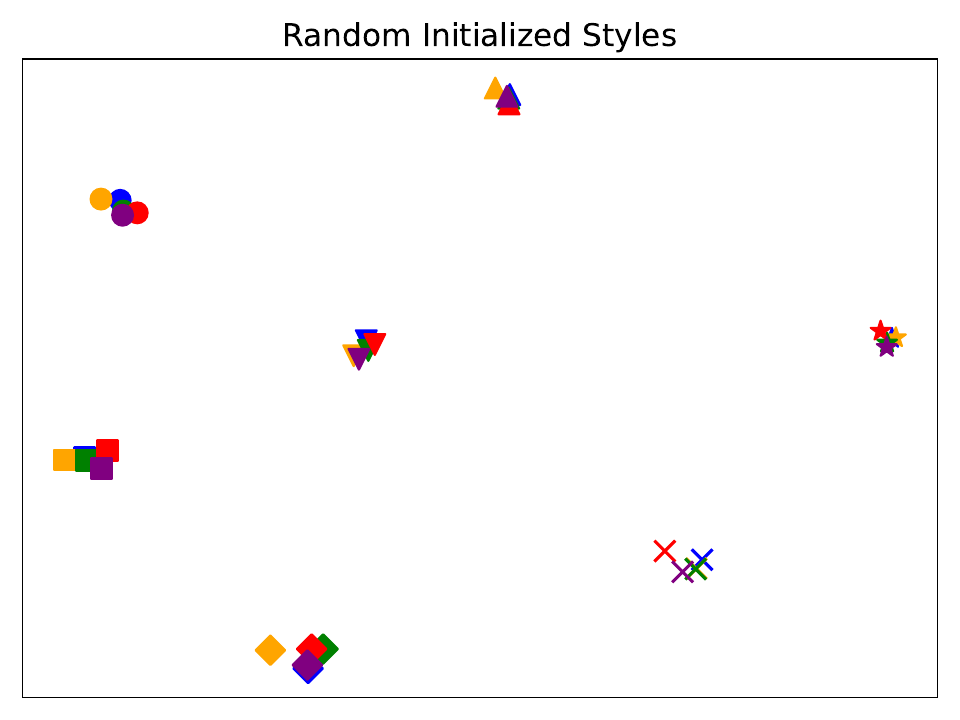}&
			\includegraphics[width=0.23\linewidth]{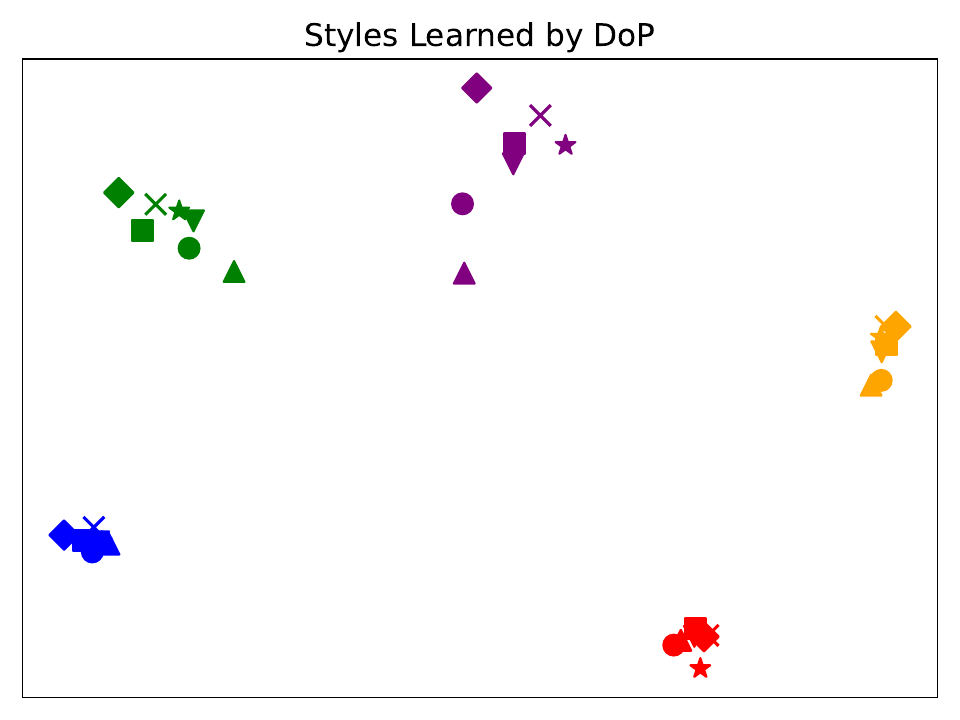}&
			\includegraphics[width=0.23\linewidth]{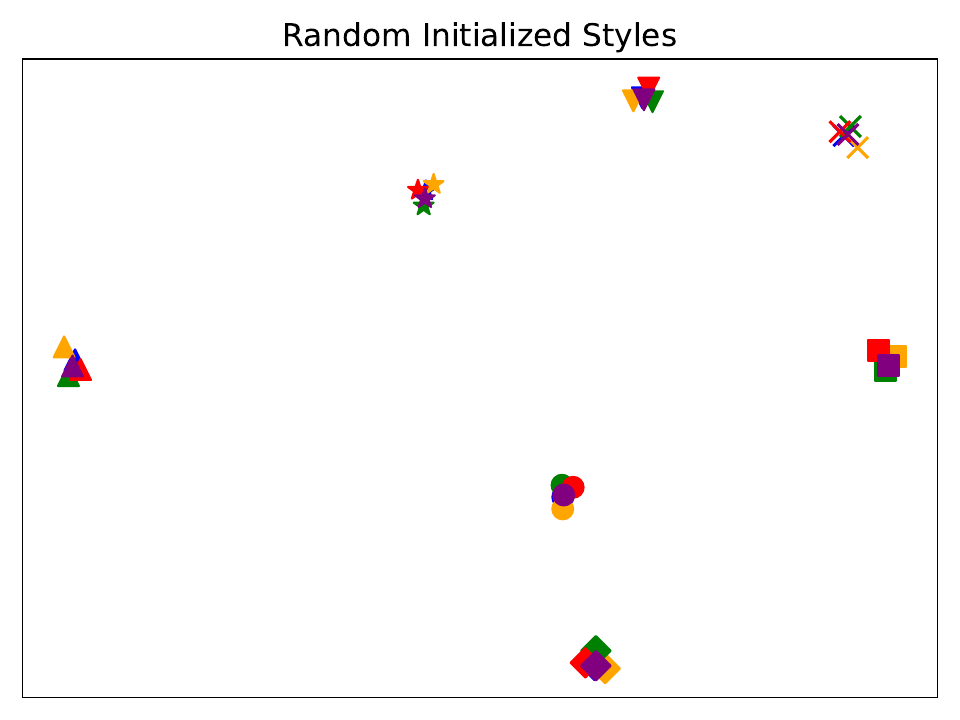}&
			\includegraphics[width=0.23\linewidth]{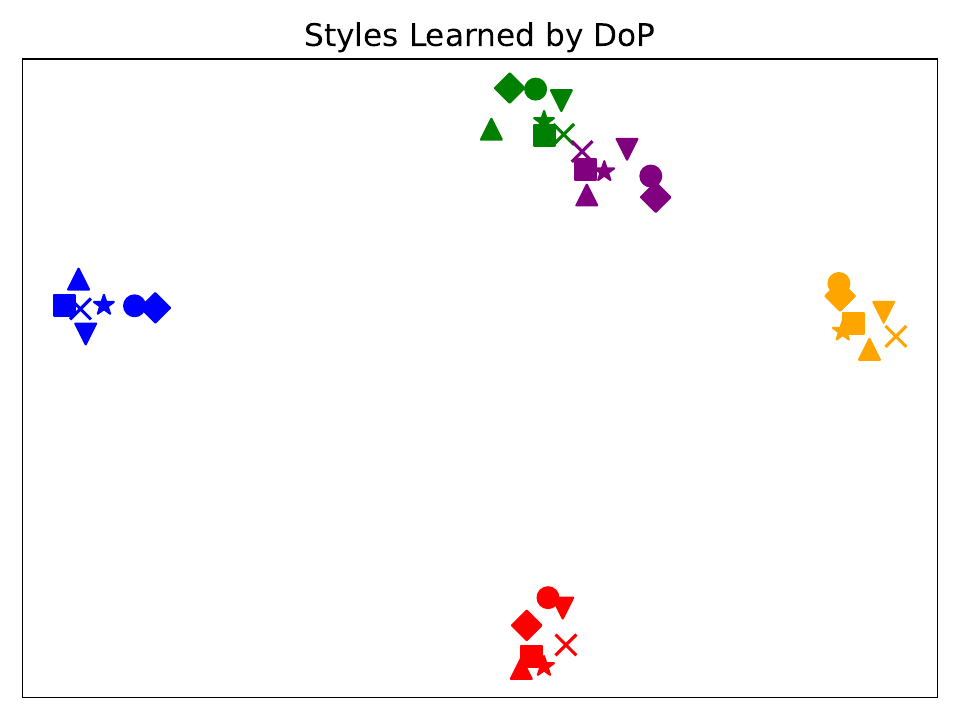}\\
			(a)  PACS  & (b) PACS & (c) OfficeHome  &  (d) OfficeHome
		\end{tabular}\vspace{-3mm}
		\caption{t-SNE visualizations of text samples with random initialized style tokens and style tokens learned by TDL of our DoP. We randomly select five styles (denoted by various colors) from $\{S_k\}_{k=1}^{10}$ and seven classes (denoted by various shapes) for easy viewing on PACS and OfficeHome datasets. Results show that DoP effectively learns diverse styles.}
		\label{fig:tsne}
	\end{center}       
\end{figure*}

\begin{figure*}[h]
	\includegraphics[width=1\linewidth]{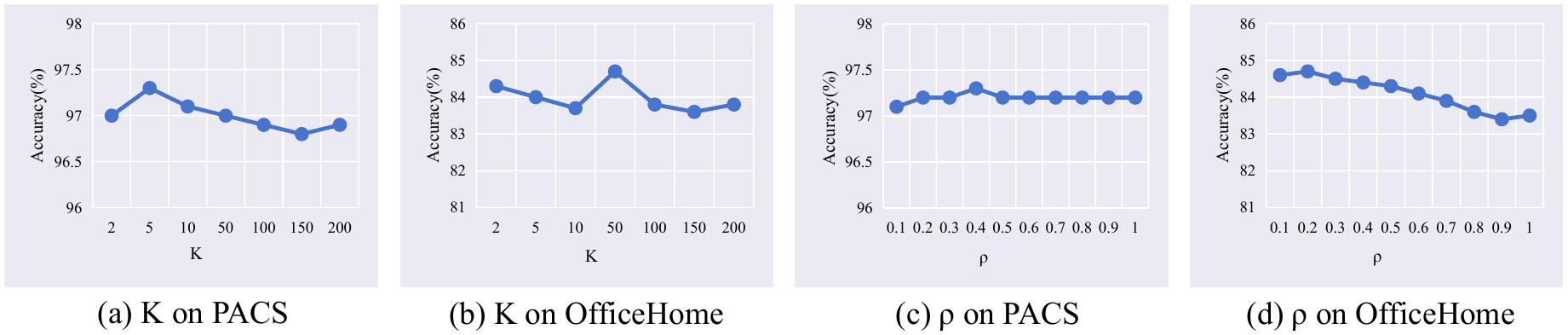}\vspace{-3mm}
	\caption{The effectiveness of the number of initialized style tokens $K$ and  confidence selection percentile $\rho$ on PACS (a, c) and OfficeHome (b, d).}
	\label{fig:initilized_num}
\end{figure*}
\subsection{Ablation Study}\label{sec4.2}
To provide further insight into DoP, we conduct ablation studies to evaluate the effects of different model components. Results on four {widely used} DG datasets are presented in  Table \ref{Table:ablation-components}.  If  DoP only takes TDL (i.e., DoP w/o VDL) for implicitly discovering domain knowledge, its average performance decreases by 6.2\%. When DoP only adopts the VDL (i.e., DoP w/o TDL) to learn domain knowledge for the target domain, its average performance decreases by 4.9\%.  Because they failed in the distribution shift between text and visual modalities in the target domain. 
{By jointly learning domain knowledge in} the text and visual spaces, {DoP improves the average accuracy} by 5.6\% and 4.3\% over DoP w/o VDL and DoP w/o TDL, respectively.
Considering the domain knowledge learned in TDL without any information supervision in the target domain, we can select the target-domain-relevant style tokens via style token selection to enable effective refinements in VDL and improve DoP (i.e., 0.6\% average performance gains). 

\subsection{Qualitative Results}\label{sec4.3}
\noindent\textbf{t-SNE Visualization of Text Samples in TDL.}  As shown in Fig. \ref{fig:tsne}, we presented the t-SNE visualization \cite{Maaten2008VisualizingDU} of text samples with randomly initialized style tokens and styles tokens learned by TDL of DoP on PACS and OfficeHome datasets. Results show that the initialized style token without obvious style means for text samples, which are clustered together according to the content information (e.g., class). In contrast, our DoP effectively compacts the content information of each text sample using the intra-style consistency loss. Accordingly, all text samples can be clustered by style. Furthermore, we can find that all the learned styles are diverse because the inter-style diverse loss encourages DoP to learn the orthogonal styles. As such, DoP effectively learns diverse styles to {provide diverse candidate style anchors for potential} target domains in text space and achieve impressive domain generalization over current state-of-the-art methods.

\noindent\textbf{ Synthesize Images According to Domain prompts.}  
To {qualitatively illustrate} the domain knowledge learned by DoP, we {use Stable Diffusion} (e.g., SD2.1 \cite{abs-2204-06125}) to synthesize {images from the learned} domain prompts with randomly initialized style tokens (i.e., $\{S_1, S_2, \cdots, S_K\}$) and style tokens learned by DoP (i.e., $S^*$), as shown in Fig. \ref{fig:style-visualization}. Results show that the style tokens learned by DoP can accurately capture the domain knowledge of target domains, while the domain prompt with random initialized style tokens failed. 
{The learned prompts capture representative characteristics of the four PACS domains, including} painting, cartoon, photo, and sketch styles.
This demonstrates DoP significantly discovers diverse target domain knowledge both in text and visual modalities, which well alleviates the challenge of domain shift in text-visual modalities. To this end, DoP consistently achieves impressive {performance} on various DG datasets for ZSDG tasks.

\subsection{Hyperparameter Analysis}\label{sec4.5}
\noindent\textbf{Number of the Initialized Style Tokens $K$.} We study the number of initialized style tokens $K$ in Eq. \ref{l_diversity} to determine its effectiveness on our DoP. As can be seen from Fig. \ref{fig:initilized_num}(a)(b), when $K$ is set to too small or too large, the performance of DoP {decreases} slightly. These results are explained as follows:  i) when $K$ is set to too small, i.e., $K$ is smaller than the numbers of target domains, the styles token cannot well cover all possible target domains; ii) when $K$ is set to too large, e.g., $K=200$ on PACS, the limited text samples may not sufficiently learn too many style tokens. Accordingly, we experimentally set the $K=10$, $K=2$, $K=50$,  $K=10$, and $K=10$ for PACS, VLCS, OfficeHome,  DomainNet,  and ImageNet, respectively. Notably, although VLCS has four domains, its domains are the “\text{photo}” style.

\noindent\textbf{Confidence Selection Percentile $\rho$.} We analyze the confidence selection percentile $\rho$  in Eq. \ref{eq:filter} to determine its effectiveness on DoP. As shown in Fig. \ref{fig:initilized_num}(c)(d), the classification accuracy is insensitive to $\rho$. This demonstrates that DoP is robust to the {choice of the} confidence selection percentile {because TDL provides well-initialized style tokens that can be stably refined by} VDL. According to the experimental results, we experimentally set the $\rho=0.1/0.1/0.2/0.1/0.1$ for PACS/VLCS/OfficeHome/DomainNet/ImageNet, respectively.

\subsection{Computational Efficiency}\label{sec4.6}
{Table~\ref{table:computational-efficiency} compares DoP with ZS-CLIP~\cite{RadfordKHRGASAM21}, PromptStyler~\cite{ChoNKYK23}, and TPT~\cite{ShuNHYGAX22} on all 9,991 PACS images using CLIP ViT-B/16. Although VDL introduces additional test-time overhead, DoP requires 170.5 ms per image and reduces the testing time by 63.2\% compared with TPT, since it refines only the selected style token $S^*$. Different from TPT which optimizes a large set of prompt parameters at test time, our token‑selective update strategy largely cuts down redundant computation. Such efficiency gain enables our method to be more practical for real‑world deployment scenarios}.

\begin{table}[!h]
	\small
	\centering
	\caption{Computational efficiency comparison on the complete PACS benchmark using CLIP ViT-B/16.
	``--'' indicates that no additional training is required.}
	\label{table:computational-efficiency}

	\begin{tabularx}{0.9\linewidth}{
		l|
		*{3}{>{\centering\arraybackslash}X}
	}
		\toprule
		Method
		& Training Time (s) $\downarrow$
		& Testing Time (s) $\downarrow$
		& Per Image (ms) $\downarrow$ \\
		\midrule

		ZS-CLIP
		& -- & 61.2 & 6.1 \\

		TPT
		& -- & 4,624.1 & 462.8 \\

		PromptStyler
		& 125.7 & 70.7 & 7.1 \\

		\textbf{DoP (Ours)}
		& 4.8 & 1,703.5 & 170.5 \\
		\bottomrule
	\end{tabularx}
\end{table}

\section{Conclusion}\label{sec6}
This paper proposes {DoP} for ZSDG, {a practical DG setting in which models recognize unseen classes in unseen domains without access to source-domain data}. The key challenge in ZSDG is the domain shift in text-visual modalities. 
{Leveraging} the strong zero-shot {capability of pretrained VLMs} (e.g., CLIP), DoP learns domain knowledge {in both the text and visual modalities to improve cross-domain generalization}.
Specifically, DoP employs TDL and VDL to learn the domain knowledge both in text and visual modalities via a prompt. The quantitative and qualitative results on four widely used DG benchmarks show that DoP effectively learns domain knowledge to perform impressive zero-shot learning and domain generalization for the ZSDG task. {These results establish} domain prompt learning {as a promising research direction} for improving the generalization of machine learning models.

Though the performance is excellent on the simple DG datasets, the results on the challenging datasets (e.g., DomainNet and ImageNet with more domains and classes) have great rest improvements. This {limitation may arise because} i) the {constructed text space contains limited domain knowledge}, and ii) a single test image {may provide insufficient information for accurately discovering target-domain characteristics}. Accordingly, we hope the empirical findings and research insights in this work could pave the way for future research on discovering rich domain knowledge from large-language models (LLM) and efficient adaptation methods for domain prompt learning.

\section*{Declarations}\label{sec7}
\textbf{Availability of Data and Materials:}
The {experiments} in this manuscript {use eight publicly available} benchmark datasets, i.e., 
PACS\footnote{\url{https://domaingeneralization.github.io/\#data}},
VLCS\footnote{\url{https://github.com/belaalb/G2DM\#download-vlcs}},
OfficeHome\footnote{\url{https://www.hemanthdv.org/officeHomeDataset.html}},
DomainNet\footnote{\url{http://ai.bu.edu/M3SDA/\#dataset}},
ImageNet-A\footnote{\url{https://github.com/hendrycks/natural-adv-examples}},
ImageNet-R\footnote{\url{https://github.com/hendrycks/imagenet-r}},
ImageNet-S\footnote{\url{https://github.com/HaohanWang/ImageNet-Sketch}},
and ImageNet-V2\footnote{\url{https://github.com/CeMOS-IS/Robust-Minisets}}, which are publicly opened benchmarks.

\noindent \textbf{Competing Interests:}
The authors declare no competing interests.

\noindent \textbf{Funding:} N/A

\noindent \textbf{Authors' Contributions:} 
Shiming Chen writes this manuscript and conduct experiments. Shuhuang Chen supports the coding.  Salman Khan and  Fahad Shahbaz Khan are the supervisors for this project and polish manuscript.

\noindent \textbf{Acknowledgements:} N/A


\bibliographystyle{spbasic}      
\bibliography{DoP}   


\end{document}